%% file: main.tex
\documentclass[10pt,twocolumn,letterpaper]{article}

\usepackage[pagenumbers]{cvpr} 


\definecolor{cvprblue}{rgb}{0.21,0.49,0.74}
\definecolor{darkgreen}{rgb}{0.0, 0.5, 0.0}
\definecolor{darkred}{rgb}{0.7, 0.0, 0.0}
\definecolor{darkblue}{rgb}{0.0, 0.0, 0.7}
\definecolor{deeporange}{RGB}{255, 69, 0}
\usepackage[pagebackref,breaklinks,colorlinks,allcolors=cvprblue]{hyperref}

\usepackage{tabularx}
\usepackage{multirow}
\usepackage{textcomp}
\usepackage{adjustbox}
\usepackage{listings}

\def\paperID{4478} 
\def\confName{CVPR}
\def\confYear{2025}

\title{VoCo-LLaMA: Towards Vision Compression with Large Language Models}

\author{
Xubing Ye$^{1}$, Yukang Gan$^{2}$, Xiaoke Huang$^{3}$, Yixiao Ge$^{2*}$, Yansong Tang$^{1*}$ \\
$^{1}$Tsinghua Shenzhen International Graduate School, Tsinghua University\\
$^{2}$ARC Lab, Tencent PCG $\quad^{3}$UC Santa Cruz\\
{\tt\small \{yxb23@mails.,tang.yansong@sz.\}tsinghua.edu.cn}  \\
{\tt\small \{brucegan,yixiaoge\}@tencent.com  \quad{xhuan192@ucsc.edu}}  \\
}

\begin{document}
\maketitle

\renewcommand{\thefootnote}{}
\footnote{Work was done when the author interned at ARC Lab, Tencent PCG. $^{*}$Corresponding author.}
\renewcommand{\thefootnote}{\arabic{footnote}} 

\input{sec/0_abstract}

\input{sec/1_Intro}
\input{sec/2_Related}

\input{sec/3_Method}

\input{sec/4_Experiments}
\input{sec/5_Conclusion}
{
    \small
    \bibliographystyle{ieeenat_fullname}
    \bibliography{main}
}


\end{document}

%% file: sec/0_abstract.tex
\begin{abstract}
Vision-Language Models (VLMs) have achieved remarkable success in various multi-modal tasks, but they are often bottlenecked by the limited context window and high computational cost of processing high-resolution image inputs and videos. 
Vision compression can alleviate this problem by reducing the vision token count.
Previous approaches compress vision tokens with external modules and force LLMs to understand the compressed ones, leading to visual information loss.
However, the LLMs' understanding paradigm of vision tokens is not fully utilised in the compression learning process.
We propose VoCo-LLaMA, the first approach to compress vision tokens using LLMs. 
By introducing \textbf{V}isi\textbf{o}n \textbf{Co}mpression tokens during the vision instruction tuning phase and leveraging attention distillation, our method distill how LLMs comprehend vision tokens into their processing of VoCo tokens.
VoCo-LLaMA facilitates effective vision compression and improves the computational efficiency during the inference stage.
Specifically, our method can achieve a 576$\times$ compression rate while maintaining 83.7\% performance. 
Furthermore, through continuous training using time-series compressed token sequences of video frames, VoCo-LLaMA demonstrates the ability to understand temporal correlations, outperforming previous methods on popular video question-answering benchmarks.
Our approach presents a promising way to unlock the full potential of VLMs' contextual window, enabling more scalable multi-modal applications.
\end{abstract}

%% file: sec/1_Intro.tex
\section{Introduction}

\begin{figure}[t]
\centering
\includegraphics[width=1\columnwidth]{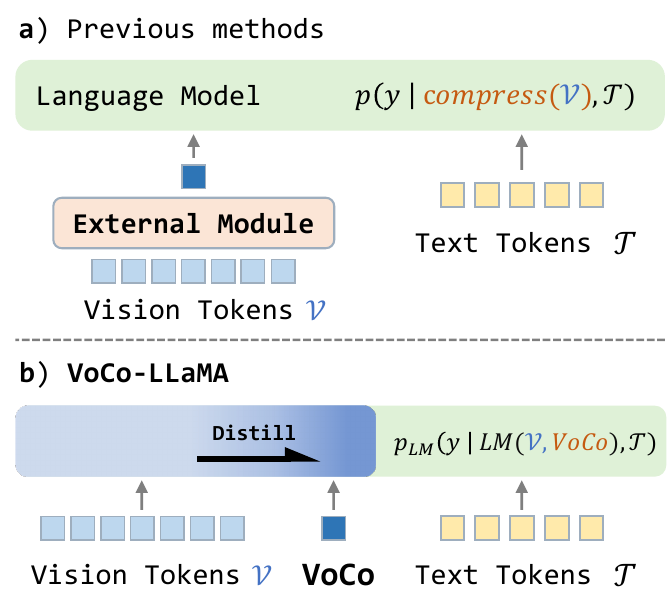}
\caption{
(a) Previous methods exploit external module, such as Q-Former~\cite{li2023blip2} or average pooling~\cite{li2023llamavid}, to ``compress'' vision tokens with substantial loss.
(b) Illustration of VoCo-LLaMA, which empowers LLM to compress vision tokens and understand compressed tokens via intrinsic token distillation.
}
\label{fig:i1}
\end{figure}

The advent of visual-language models~\cite{li2023blip2, liu2023llava, zhu2023minigpt, du2022glm, Qwen-VL, internlmxcomposer, liu2023improvedllava, fuyu2023, liu2023world, geminiteam2024gemini} has led to significant advancements in visual understanding. 
Particularly, high-resolution image encoding~\cite{liu2023improvedllava, fuyu2023} and the incorporation of more video frames~\cite{liu2023world, geminiteam2024gemini} have been shown to enhance the capabilities of both large visual-language models and large video-language models, respectively. 
However, the large number of vision tokens occupies a substantial portion of the valuable context window of the large language model, leading to expensive computational costs.
For instance, when using high-resolution image inputs in LLaVA-1.6~\cite{liu2023improvedllava}, a single image with a resolution of 672 $\times$ 672 is divided into smaller patches, each encoded with a 336 $\times$ 336 resolution input.
This process yields an image representation consisting of 2880 vision tokens, occupying over half of the context length.
As the number of input images increases, the context window for text will be further bottlenecked.
\cite{liu2023world, geminiteam2024gemini} investigate the efficacy of extending the context length to the million-level mark to mitigate this issue, but this approach necessitates expensive computational resources (\emph{e.g.,} \cite{liu2023world} requires over 1000 v4 TPUs) and engineering efforts in data and framework development.

To address this issue, previous methods~\cite{li2023blip2, zhu2023minigpt, du2022glm, internlmxcomposer, li2023llamavid, instructblip} have exploited Q-Former~\cite{li2023blip2} or Re-sampler~\cite{alayrac2022flamingo} to ``compress'' the encoded vision tokens.
As illustrated in ~\cref{fig:i1} (a), these kind of methods compress the vision tokens with \textbf{external} modules and force LLMs to understand the compressed ones.

Given that the LLM can effectively understand uncompressed vision tokens~\cite{liu2024visual}, it has great potential to perform token compression on its own. Therefore, we propose \textbf{VoCo-LLaMA}, 
the first vision compression method that leverages the inherent capabilities of large language models to our best knowledge.
As demonstrated in ~\cref{fig:i1} (b), we introduce \textbf{V}isi\textbf{o}n \textbf{Co}mpression (\textbf{VoCo}) tokens between visual and text tokens. 
By modifying the attention mechanism, we ensure that VoCo tokens attend exclusively to visual tokens, while text tokens attend solely to VoCo tokens.
Subsequently, we establish an exclusive interaction pathway between the visual and text tokens via VoCo tokens.
This facilitates the LLM itself to compress and distill the parsing vision tokens, specifically the transformer activations on top of them, into compact VoCo tokens.

Building upon this, we further investigate the efficacy of VoCo-LLaMA in handling video input. The total number of visual tokens for each video can be substantial, far exceeding the context length of large language models (LLMs), making it impractical to utilize VoCo-LLaMA to compress all the tokens simultaneously. To address this issue, we first employ VoCo-LLaMA to compress the visual tokens of each frame into voco tokens. These voco tokens are subsequently concatenated to form a sequential token series. VoCo-LLaMA then extracts both visual and temporal information from this series to facilitate video understanding tasks. With this effective design, the LLMs can handle much longer videos within the same context length.

During inference, VoCo-LLaMA mitigates the issue of limited context length in LLM by employing a two-stage forward process. The fist stage compresses visual tokens of each image into a reduced set of VoCo tokens, while the second stage completes the task by utilizing both VoCo tokens and text tokens. 
Moreover, VoCo tokens can be cached and reused when handling various tasks involving identical visual inputs, thereby enhancing computational efficiency and reducing storage requirements compared to maintaining the entire KV-cache for uncompressed visual tokens.
Experimental results on various benchmarks demonstrate that VoCo-LLaMA achieves a 576x compression rate while maintaining 83.7\% of the original performance. Additionally, significant reductions in inference computation costs were observed, including up to 99.8\% in cache storage, 94.8\% in FLOPs, and 69.6\% in inference time.

Our core contributions are summarized as follows:
\begin{itemize}
    \item We propose VoCo-LLaMA, the first approach to compress vision tokens by leveraging the inherent capabilities of large language models, thereby eliminating the need for any external modules.
    \item We extend VoCo-LLaMA from image input to video input, which allows the LLM to handle approximately 200 times more video frames while maintaining its video understanding capabilities.
    \item Extensive experiments on image and video benchmarks demonstrate the effectiveness of our method, showcasing superior performance in both token compression and inference efficiency compare to various existing baselines.
\end{itemize}

%% file: sec/2_Related.tex
\section{Related Work}

\textbf{LLMs and Text Compression.}
In recent years, large language models (LLMs) have sparked a technological revolution. 
As the scale of training data and model size continue to expand, models~\cite{vicuna2023, gpt3, gpt4, workshop2023bloom, touvron2023llama, Hugo2023, jiang2023mistral, huang2024good} have demonstrated exceptional capabilities in understanding and generating language.
In particular, models such as the LLaMA series~\cite{touvron2023llama, Hugo2023, huang2024good, vicuna2023} 
have emerged as foundational models or main components in many research works. 
However, the limited context window size in LLMs has long been a widely discussed topic.
Text compression has been proven to be an efficient approach. 
Long-standing works, including~\cite{dai-etal-2019-transformer, liu2018generating, Rae2019CompressiveTF, wu2022memorizing, Zhang2021PoolingformerLD}, focus on storing text representations in transformers to achieve dense information representation.
\cite{Askell2021AGL, Snell2022LearningBD} have demonstrated the effectiveness of distilling long text information into prompt-free student models. 
In a similar vein, recent studies~\cite{wingate2022promptcompressioncontrastiveconditioning, chevalier2023adapting, mu2024learning, ge2024incontext} have explored the potential applications of compressing text in large language models. 
However, the discussion of compressing visual information has been relatively understudied compared to the language model domain. 
Our work pioneers the use of LLMs' learning capabilities to compress vision information, aiming to bridging this gap in the field of VLMs.

\begin{figure*}[t]
\centering
\includegraphics[width=0.94\linewidth]{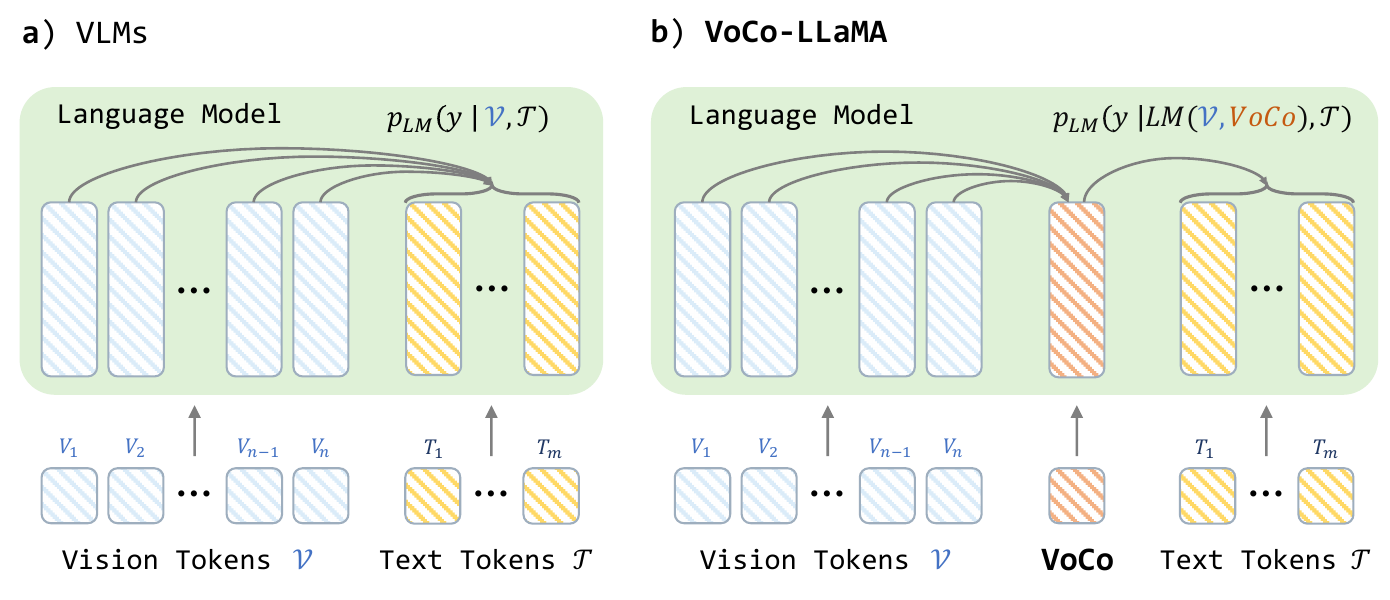}
\caption{
Illustration of the VoCo-LLaMA framework. 
Based on standard VLMs (a), VoCo-LLaMA (b) first isolate visual and text tokens by injecting VoCo tokens, and then establishes a dedicated interaction pathway between the two modalities via VoCo tokens, enabling effective compression of vision tokens into the transformer activations upon the compact VoCo tokens.
}
\label{fig:m1}
\end{figure*}

\noindent\textbf{VLMs and Vision Compression.}
The success of LLMs has inspired significant progress in vision language models (VLMs). 
By integrating visual encoders with LLMs, VLMs can effectively achieve cross-modal understanding through instruction tuning.
Previous methods~\cite{li2023blip2, liu2023llava, zhu2023minigpt, du2022glm, Qwen-VL, internlmxcomposer, liu2023improvedllava, fuyu2023, instructblip, alayrac2022flamingo} have substantiated the success of this training paradigm in visual understanding. 
The successful application of VLMs on images has also been rapidly extended to the video domain~\cite{liu2023world, geminiteam2024gemini, li2023llamavid, huang2023vtimellm, jin2024chatunivi, damonlpsg2023videollama, 2023videochat, liu2023btadapter, luo2023valley, Maaz2023VideoChatGPT}. 
With the input of higher-resolution images~\cite{liu2023improvedllava, fuyu2023} and more video frames~\cite{liu2023world, geminiteam2024gemini}, VLMs can capture rich visual information.
However, as the number of vision tokens representing an input image increases, they take up a significant portion of the limited context window of language models, and can even exceed it.
To address this, previous methods~\cite{li2023blip2, zhu2023minigpt, du2022glm, internlmxcomposer, instructblip} have largely employed Q-Former~\cite{li2023blip2}, which maps images to fixed-length tokens in the language embedding space through learnable queries, compressing visual information. 
A more recent approach~\cite{li2023llamavid} has applied average pooling with a learnable linear layer to compress visual features through multi-stage training strategy. 
Although these methods perform moderately well at lower compression multiples, they cause a significant loss of valuable visual information when the number of compressed tokens reduces.
VoCo-LLaMA distills the approach of LLMs in understanding vision tokens into their processing of compressed tokens, significantly reducing information loss during the vision compression process.

%% file: sec/3_Method.tex
\section{Method} \label{sec:method}

We first introduce VoCo-LLaMA, a large language model capable of compressing lengthy vision tokens into compact VoCo tokens through attention distillation, which enables the efficient representation of visual information.
Then, we build upon these compressed tokens to continue training VoCo-LLaMA, enabling our model to capture temporal dependencies within video data.

\subsection{Vision Compression} \label{sec:vision compression}

Given a paired image and text input, we follow the design of most vision-language models (VLMs) and encode the image into a sequence of vision tokens $\mathcal{V} = \{V_1, \ldots, V_n\}$, where $n$ is the number of the output patches from the visual encoder. 
Similarly, the text input is encoded into a sequence of text tokens $\mathcal{T} = \{T_1, \ldots, T_m\}$. 
Consider an original, unmodified standard large vision language model (denoted as $LM_o$), exemplified by LLaVA~\cite{liu2023llava}, depicted in~\cref{fig:m1} (a).
During visual instruction tuning, $LM_o$ leverages both vision tokens $\mathcal{V}$ and text tokens $\mathcal{T}$ to predict the output $y$, and learns the distribution $p_{LM_o}(y\mid\mathcal{V}, \mathcal{T})$. 
For image compression models, our goal is to employ a compact set of compressed tokens $\mathcal{C}$ to efficiently represent the vision token set $\mathcal{V}$. 
Additionally, we aim to generate outputs that closely approximates the outputs of the original model $LM_o$ when presented with identical image and text inputs.

With an image encoded as vision tokens $\mathcal{V}$, we formulate the image compression distillation process as learning a compression model $LM_{c}$ that generates the output $y$ conditioned on the compressed tokens $\mathcal{C}$ and the text tokens $\mathcal{T}$. 
This is achieved by learning a probability distribution $p_{LM_{c}}(y\mid\mathcal{C}, \mathcal{T})$. 
The optimization objective of $LM_{c}$ is to minimize the loss function:
\begin{align}
  E_{\mathcal{V},\mathcal{T}}[D_{KL}(p_{LM_o}(y\mid\mathcal{V}, \mathcal{T}) \parallel p_{LM_{c}}(y\mid\mathcal{C}, \mathcal{T}))]
  \label{eq:1}
\end{align}

With above distillation objective, how to further distill the information within the vision tokens $\mathcal{V}$ into the compressed token $\mathcal{C}$ is the key of vision compression.

\begin{figure*}[t]
\centering
\includegraphics[width=1\linewidth]{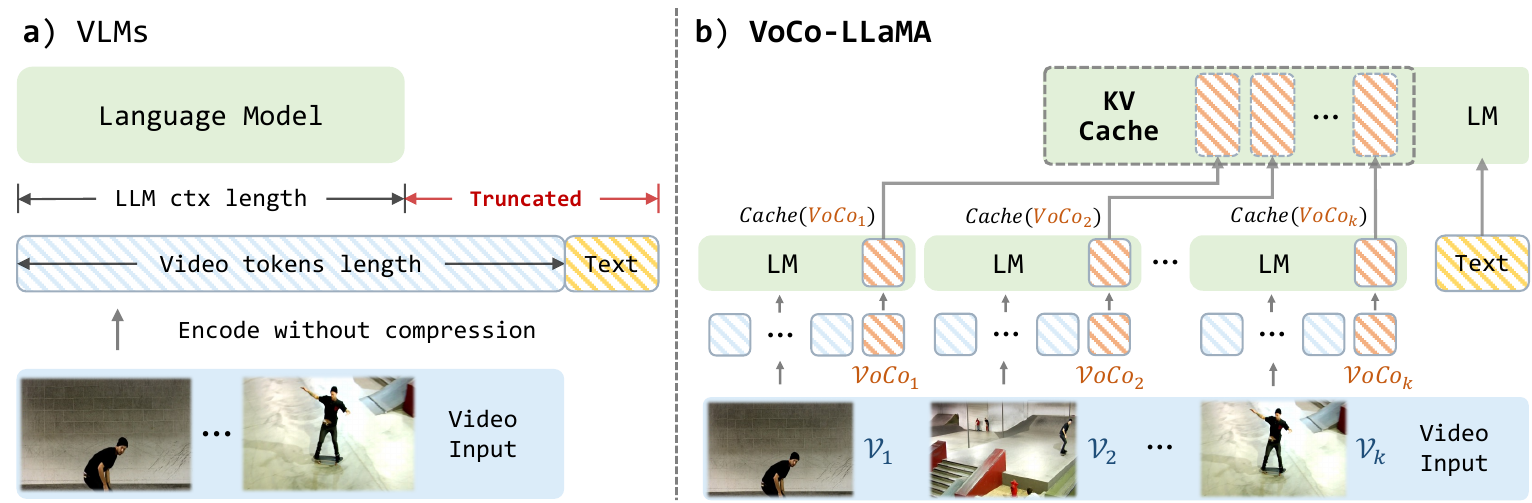}
\caption{
(a) VLMs are bottlenecked by the limited context window when processing video frames. 
(b) Extension of VoCo-LLaMA to video domain: Enabling more frames input with a limited context length.
}
\label{fig:m2}
\end{figure*}

\subsection{VoCo-LLaMA} \label{sec:voco-llama}
As illustrated in~\cref{fig:m1} (b), VoCo-LLaMA leverages the LLM's ability to compress visual tokens into compact \textbf{V}isi\textbf{o}n \textbf{Co}mpression (VoCo) tokens and learns to understand image through these VoCo tokens.
The input sequence to the large language model is formed by concatenating the vision tokens, the special $VoCo$ tokens, and the text tokens, yielding a sequence:
\begin{align} 
\label{eq:2}
(\mathcal{V}, VoCo, \mathcal{T}) = (V_0, \ldots, V_n, VoCo, T_0, \ldots, T_m)
\end{align}
In the training phase, we employ a two-stage attention mechanism.
Initially, we impose a constraint on the text tokens, explicitly preventing them from attending to the original vision tokens, and requiring them to exclusively attend to the compressed and distilled VoCo tokens.
Subsequently, the vision tokens are subjected to continuous attention from the VoCo tokens due to the casual attention mechanism. 
This deliberate design ensures that the text tokens solely capture the distilled visual information encoded in the VoCo tokens, rather than directly interacting with the original vision tokens, thereby achieving effective compression from vision tokens to compressed tokens.

The compression process of VoCo-LLaMA can be elegantly implemented by modifying the attention mask.
Specifically, we set the attention weights between the text tokens and the vision tokens to $False$, effectively rendering the text tokens ``\textbf{isolated}" to the vision tokens.
Formally, let $\mathbf{M} \in \mathbb{R}^{(m+n+1) \times (m+n+1)}$ denote the attention mask, where $M_{ij} = True$ if token $i$ attends to token $j$, and $M_{ij} = False$ otherwise. We define the attention mask as:
\begin{align} 
\label{eq:3}
M_{ij} = \begin{cases}
True, & \text{if } i \in \mathcal{T} \text{ and } j \in VoCo, \\
False, & \text{if } i \in \mathcal{T} \text{ and } j \in \mathcal{V}, \\
True, & \text{otherwise}.
\end{cases}
\end{align}

In practice, VoCo-LLaMA can be effectively trained under the standard supervised fine-tuning paradigm, leveraging the abundant image-text data readily available in VLMs. 
Furthermore, the VoCo token can be compactly represented as a set of Transformer activations, allowing them to be cached to enhance inference efficiency, which will be discussed in~\cref{sec:cache}.

VoCo-LLaMA enables the large language models to learn the compression process of vision tokens, $LM(\mathcal{V}, VoCo)$, while simultaneously learning to understand the compressed VoCo tokens. We define the target learning distribution as follows:
\begin{align}
\label{eq:4}
p_{VoCo-LLaMA} = p_{LM}(y \mid LM(\mathcal{V}, VoCo), \mathcal{T})
\end{align}
the optimization objective in~\cref{eq:1} can be defined as:
\begin{align}
\label{eq:5}
E_{\mathcal{V},\mathcal{T}}[D_{KL}(p_{LM_o}(y\mid\mathcal{V}, \mathcal{T}) \mid\mid p_{VoCo-LLaMA}]
\end{align}

\begin{figure}[t]
\centering
\includegraphics[width=1\columnwidth]{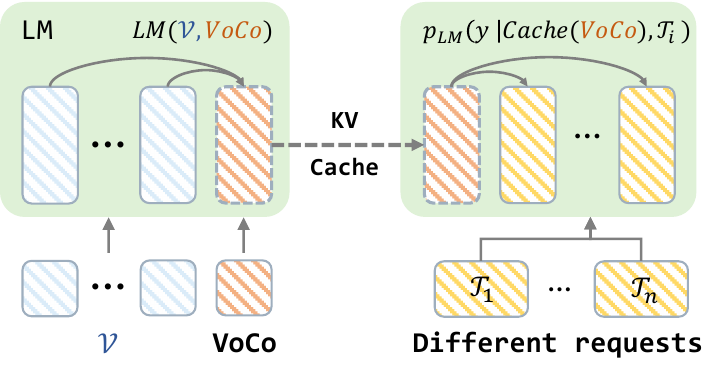}
\vspace{-15pt} 
\caption{
Illustration of the two stage forward operation with KV cache for VoCo-LLaMA during inference.
The first forward pass extract image into VoCo cache.
The cached VoCo tokens can be utilized to handle different taksk that involve same image.
}
\label{fig:m3}
\end{figure}

\subsection{Reuse of VoCo Cache} \label{sec:cache}
During inference, VoCo-LLaMA mitigates the issue of limited context window size by dividing the single forward pass into two phases.
As illustrated in~\cref{fig:m3}, the first forward pass takes [vision tokens, VoCo tokens] as input to compress visual information into Transformer activations upon VoCo tokens.
The second forward pass takes [text tokens] as input and load VoCo activations as KV Cache.

Moreover, VoCo tokens derived from the first forward pass can be cached and reused when handling various tasks involving identical image/video inputs, thereby enhancing computational efficiency and reducing storage requirements compared to maintaining the entire KV-cache for uncompressed visual tokens.
For more details on the inference implementation, please refer to the \textit{supplementary material}.

\subsection{Temporal Modeling} \label{sec:temporal modeling}
When giving a sequence of video frames $Vid = \{\mathcal{V}_1, \ldots, \mathcal{V}_k\}$ and a corresponding text input, the token length for the entire video far exceeds LLM context length, as shown in~\cref{fig:m2} (a).
To solve this issue, VoCo-LLaMA divides video input into smaller segments and input these segments into LLM with VoCo tokens $\{VoCo_1, \ldots, VoCo_k\}$. 
As shown in~\cref{fig:m2} (b), all the frames are compressed into VoCo activations.
Specifically, we obtain the compressed representation $Cache_t$ for video segment tokens $\mathcal{V}_t$ through $Cache(VoCo_t) = LM(\mathcal{V}_t, VoCo_t)$.
This yields a sequence of KV Cache representing compressed video tokens, denoted by $\mathcal{F} = \{Cache(VoCo_1), \ldots, Cache(VoCo_k)\}$.

Having obtained the time-series compressed cache sequences $\mathcal{F}$, we enable language model to capture and comprehend the temporal correlations among the compressed video tokens. 
With the inclusion of text tokens $\mathcal{T}$, VoCo-LLaMA learns the distribution $p(y \mid \mathcal{F}, \mathcal{T})$.
We adopt a continue training process based on VoCo-LLaMA with image compression capabilities, allows the model to focus on temporal modeling, thereby streamlining the video understanding process.

\subsection{Implementation Details} \label{sec:implementation details}

Regarding the training strategy and data, as mentioned earlier in~\cref{sec:voco-llama}, VoCo-LLaMA only requires learning to insert and compress VoCo tokens during the vision instruction tuning stage.
We follow the common VLMs~\cite{liu2023llava, liu2023improvedllava} to encode the image input into vision tokens with vision encoder and a linear projector.
We employ the pre-trained CLIP-ViT-L~\cite{Radford2021LearningTV} as our visual encoder.
For pre-trained large language models, we utilize Vicuna-7B~\cite{vicuna2023}. 
Without introducing VoCo tokens, we first align the visual encoder and language model using the LLaVA-filtered CC3M~\cite{sharma2018conceptual} dataset with visual encoder and language model keeping frozen. 
During the instruction tuning phase of VoCo-LLaMA, incorporating multiple image understanding tasks is crucial for learning a scalable image compression model. 
Therefore, we construct the instruction pairs inspired by~\cite{li2023llamavid} using~\cite{liu2023improvedllava}.
For video tuning, we further utilize WebVid~\cite{Bain21} and the QA-pairs of Video-ChatGPT~\cite{Maaz2023VideoChatGPT}.
Moreover, gradient checkpointing strategies are employed to reduce computational cost during training.

We conducted experiments on several common compression strategies with the same training setting and data for comparison.
For the compression strategy with Q-Former, we employ the architecture in~\cite{li2023blip2} and configure the query number to one, resulting in a single compression token. 
For the compression strategy with average pooling, we follow the design of the single content token in~\cite{li2023llamavid}. 
For more details on the training and inference implementation, please refer to the \textit{supplementary material}.

%% file: sec/4_Experiments.tex
\section{Experiments}  \label{sec:experiments}

\renewcommand{\arraystretch}{1.0}  
\begin{table*}[t]
  \centering
  \begin{tabularx}{0.99\textwidth}{l|>{\centering\arraybackslash}X|*{8}{>{\centering\arraybackslash}X}}
    \toprule
    Method & Token & \textbf{GQA} & \textbf{MMB} & \textbf{MME} & \textbf{POPE} & \textbf{SEED} & \textbf{SQA$^{I}$} & \textbf{VQA$^{v2}$} & \textbf{Avg.} \\
    \midrule
    \multirow{2}{*}{{\textcolor{darkgreen}{Upper Bound}}} & \multirow{2}{*}{{\textcolor{darkgreen}{576}}} & {\textcolor{darkgreen}{61.1}} & {\textcolor{darkgreen}{64.0}} & {\textcolor{darkgreen}{1487.2}} & {\textcolor{darkgreen}{85.0}} & {\textcolor{darkgreen}{57.9}} & {\textcolor{darkgreen}{66.5}} & {\textcolor{darkgreen}{77.7}} & - \\
    & & \textcolor{darkgreen}{100{\footnotesize \%}}  & \textcolor{darkgreen}{100{\footnotesize \%}} & \textcolor{darkgreen}{100{\footnotesize \%}} & \textcolor{darkgreen}{100{\footnotesize \%}} & \textcolor{darkgreen}{100{\footnotesize \%}} & \textcolor{darkgreen}{100{\footnotesize \%}} & \textcolor{darkgreen}{100{\footnotesize \%}} & \textcolor{darkgreen}{100{\footnotesize \%}} \\
    
    \midrule
    \multirow{2}{*}{\textbf{VoCo-LLaMA}}  & \multirow{2}{*}{1} & \textbf{57.0} & \textbf{58.8} & \textbf{1323.3} & \textbf{81.4} & \textbf{53.7} & \textbf{65.4} & \textbf{72.3} & - \\

    & & \textbf{82.5{\footnotesize \%}} & \textbf{87.5{\footnotesize \%}} & \textbf{81.2{\footnotesize \%}} & \textbf{88.4{\footnotesize \%}} & \textbf{80.0{\footnotesize \%}} & \textbf{81.0{\footnotesize \%}} & \textbf{85.2{\footnotesize \%}} & \textbf{83.7{\footnotesize \%}} \\
    
    \midrule
    Avg. Pool~\cite{li2023llamavid}  & \multirow{2}{*}{1} & 52.9 & 55.5 & 1210.3 & 79.1 & 50.3 & 62.2 & 65.0 & -  \\ 
     $+$ Linear   & & 65.0{\footnotesize \%} & 79.6{\footnotesize \%} & 68.1{\footnotesize \%} & 81.0{\footnotesize \%} & 63.8{\footnotesize \%} & 25.8{\footnotesize \%} & 65.2{\footnotesize \%} & 64.1{\footnotesize \%}\\
     
    \midrule
    \multirow{2}{*}{Q-Former~\cite{li2023blip2}}     & \multirow{2}{*}{1} & 51.1 & 51.7 & 1079.7 & 77.3 & 47.2 & 62.7 & 63.4 & -  \\ 
     & & 57.3{\footnotesize \%} & 70.5{\footnotesize \%} & 53.2{\footnotesize \%} & 75.2{\footnotesize \%} & 49.0{\footnotesize \%} & 34.5{\footnotesize \%} & 60.8{\footnotesize \%} & 57.2{\footnotesize \%}\\
     
    \midrule
    \multirow{2}{*}{{\textcolor{darkred}{Lower Bound}}}  & \multirow{2}{*}{{\textcolor{darkred}{1}}} & {\textcolor{darkred}{37.7}} & {\textcolor{darkred}{22.3}} & {\textcolor{darkred}{617.3}} & {\textcolor{darkred}{53.9}} & {\textcolor{darkred}{36.9}} & {\textcolor{darkred}{60.7}} & {\textcolor{darkred}{41.2}} & - \\
    & & \textcolor{darkred}{0{\footnotesize \%}} & \textcolor{darkred}{0{\footnotesize \%}} & \textcolor{darkred}{0{\footnotesize \%}} & \textcolor{darkred}{0{\footnotesize \%}} & \textcolor{darkred}{0{\footnotesize \%}} & \textcolor{darkred}{0{\footnotesize \%}} & \textcolor{darkred}{0{\footnotesize \%}} & \textcolor{darkred}{0{\footnotesize \%}} \\
    \bottomrule
  \end{tabularx}
\vspace{-3pt} 
  \caption{
  Comparison with previous approaches on vision compression using common visual understanding benchmarks.
  All methods compress 576 vision tokens (from $(336/14)^2 = 576$) into one.
  We further report the compression performance mentioned in~\cref{sec:implementation details}.
}
\vspace{-7pt} 
\label{tab:voco_tab_1}
\end{table*}

\renewcommand{\arraystretch}{0.9} 
\begin{table}[t]
\small
\centering
\resizebox{0.99\columnwidth}{!}{%
\begin{tabularx}
{\linewidth}{>{\centering\arraybackslash}X|*{5}{>{\centering\arraybackslash}X}}
    \toprule
    Token & \textbf{MMB}  & \textbf{GQA} & \textbf{VQA$^{v2}$} & \textbf{SEED} & \textbf{Avg.} \\
    \midrule
     {\textcolor{darkgreen}{576}} & {\textcolor{darkgreen}{64.0}} & {\textcolor{darkgreen}{61.1}} &  {\textcolor{darkgreen}{77.7}} & {\textcolor{darkgreen}{57.9}} & {\textcolor{darkgreen}{100{\footnotesize \%}}}  \\
    \midrule
    128 & \textbf{61.0} & 59.8 & \textbf{76.9} & \textbf{59.1} & \textbf{97.7{\footnotesize \%}}    \\
    64 & 60.5 & \textbf{60.4} & 75.4 & 56.3 & 93.7{\footnotesize \%}    \\
    32 & 59.4 & 60.2 & 75.3  & 56.2  & 92.6{\footnotesize \%} \\
    16 & 58.6 & 59.4 & 75.4 & 56.2 & 91.3{\footnotesize \%}    \\
    8 & 58.7 & 59.2 & 75.3 & 56.3 & 91.3{\footnotesize \%}   \\
    4 & 60.4 & 58.4 & 74.5 & 56.0 & 90.4{\footnotesize \%}   \\
    2 & 60.1 & 57.7 & 73.5 & 55.0 & 87.8{\footnotesize \%}    \\
    1 & 58.8 & 57.0 & 72.3 & 53.7 & 83.8{\footnotesize \%}    \\
    \midrule
   {\textcolor{darkred}{1}} & {\textcolor{darkred}{22.3}} & {\textcolor{darkred}{37.7}} & {\textcolor{darkred}{41.2}} & {\textcolor{darkred}{36.9}} & {\textcolor{darkred}{0{\footnotesize \%}}}   \\
    \bottomrule
\end{tabularx}}
\vspace{-2pt} 
\caption{
Effect of VoCo tokens count on widely used benchmarks.
The number of VoCo tokens increases from 1 to 128.
{\textcolor{darkgreen}{Green}} and {\textcolor{darkred}{red}} represent the Upper and Lower Bound, respectively.
}
\vspace{-8pt} 
\label{tab:voco_tab_2}
\end{table}

\subsection{Datasets} 
In this work, we conduct experiments on several common visual understanding benchmarks for vision compression.
In particular, we report results on GQA~\cite{hudson2018gqa}, MMB (MMBench)~\cite{MMBench}, MME~\cite{fu2023mme}, POPE~\cite{Li2023pope}, SEED-Bench~\cite{li2023seed}, SQA$^I$ (Image-based setting in ScienceQA)~\cite{lu2022sqa} and VQA$^{v2}$ (VQA V2)~\cite{balanced_vqa_v2}.
By observing the model's performance on these image understanding benchmarks before and after compression (\emph{i.e.} with initial vision tokens / VoCo tokens), we can observe the effects of the visual information loss that occurs during the vision compression process.
We evaluate the performance on these visual understanding benchmarks in accordance with the details outlined in~\cite{liu2023llava}.
As for the video domain, we evaluate the zero-shot performance on several video question-answering benchmarks.
MSVD-QA~\cite{xu2017video} is a video QA dataset consisting of 1,970 video clips with 50,505 QA pairs, built upon the Microsoft Research Video Description Corpus~\cite{chen2011collecting}.
MSRVTT-QA~\cite{xu2017video} is a large-scale video QA dataset featuring 10K videos and 243K question-answering pairs with complex scenes, based on the MSR-VTT dataset~\cite{xu2016msr}. 
ActivityNet-QA~\cite{yu2019activityqa} is a fully annotated video QA dataset containing 58K question-answering pairs derived from 5,800 complex web videos from the ActivityNet dataset~\cite{Gupta2022VisProg}.

\subsection{Vision Compression Configuration}  \label{sec:compress config}

In the primary experiment of vision compression, we present the results of compressing all vision tokens of an image into a single VoCo token. 
To rigorously quantify the performance loss of VoCo-LLaMA during compression, we designed two comparative training settings: 
the \textcolor{darkgreen}{Upper Bound} model, which represents the best compression performance. The ideal case for a visual compression model is to obtain the same visual understanding capability as the upper bound model.
And the \textcolor{darkred}{Lower Bound} model, which represents the worst compression performance.

The initialization model is trained by integrating VoCo tokens in a manner analogous to VoCo-LLaMA, without modifying the attention mask strategy. 
During inference, we employ a standard causal attention mask. 
This setting effectively controls for performance fluctuations induced by the introduction of additional special tokens.
In contrast, the random compression model is trained under identical settings as the initialization model. 
During inference, we restrict the visibility of text tokens to only the VoCo token, isolating the visual information. 
This setup represents a scenario without vision compression training, providing a baseline for evaluating.
Based on the performance boundary model, the compression retention rate can be subsequently calculated as $(\text{result of VoCo-LLaMA} - \text{Lower Bound}) / (\text{Upper Bound} - \text{Lower Bound})$.

\subsection{Results} \label{sec:results}


\textbf{\hypertarget{subsubsec}{Vision Compression.}}
\cref{tab:voco_tab_1} presents the results of VoCo-LLaMA on vision compression. 
To explore the maximum potential of our method, we report the highest achievable compression ratio, which compresses vision tokens into one single VoCo token. 
We report results of our compression model on various common visual understanding benchmarks, as well as the compression retention rates defined based on upper and lower bound models introduced in~\cref{sec:compress config}.
It can be observed that our method preserves the original visual information to a large extent, even at an extremely high compression ratio of 576$\times$. 
Specifically, we achieved an average compression retention rate of 83.7\% across seven widely used benchmarks. 
Especially on MMBench, POPE and VQA$^{v2}$, our method retained more than 85\% of the performance during compression.
The results indicate that VoCo-LLaMA can effectively compress vision tokens.
Moreover, our method consistently outperforms the performance lower bound model of random compression across all benchmarks. 
This demonstrates that the advantages of VoCo-LLaMA, such as significant reductions in context length and improved calculation efficiency, outweigh the performance loss caused by compression.

\renewcommand{\arraystretch}{0.9} 
\begin{table}[t]
\small
  \centering
\resizebox{0.99\columnwidth}{!}{%
  \begin{tabular}{l|>{\centering\arraybackslash}p{0.6cm}|*{4}{c}}
    \toprule
    Method & \textit{N} & \textbf{GQA} & \textbf{POPE} & \textbf{SQA$^{I}$} & \textbf{VQA$^{T}$} \\
    \midrule
    \multirow{3}{*}{LLaMA-VID~\cite{li2023llamavid}} 
      & 16 & 58.2 & 83.1 & 67.4 & 50.8    \\
      & 4 & 56.2 & 83.5 & 68.7 & 49.1    \\
      & 1 & 55.5 & 83.1 & 68.8 & 49.0    \\
    \midrule
    \textbf{VoCo-LLaMA} & 1 & \textbf{58.3} & \textbf{85.0} & \textbf{69.5} & \textbf{52.7}    \\
    \bottomrule
  \end{tabular}
  }
\vspace{-1pt} 
  \caption{
  Comparison with previous compression methods which compress image into single token.
  $N$ means the number of ``content'' tokens in LLaMA-VID or the VoCo tokens in our method.
  The input resolution is set to 224 for fair comparison.
  }
\label{tab:voco_tab_3}
\end{table}

\renewcommand{\arraystretch}{0.9} 
\begin{table}[t]
\footnotesize
  \centering
\resizebox{0.99\columnwidth}{!}{%
    \begin{tabular}{l|c|cccc}
    \toprule
     {Method} & {Token}  & {\textbf{MMB}}& {\textbf{GQA}} & {\textbf{VQA$^{v2}$}} & {\textbf{SEED}}   \\
    \midrule
        \multirow{4}{*}{\textbf{VoCo-LLaMA}} & 32 & \textbf{59.4} & \textbf{60.2} & \textbf{75.3}  & \textbf{56.2}   \\
        & 16 & {58.3} & {58.9} & {74.9} & {55.8}    \\
       & 4 & {59.7} & {58.0} & {73.5} & {55.2}    \\
       & 1 & {57.9} & {56.1} & {71.2} & {53.0}    \\
    \bottomrule
    \end{tabular}}
\vspace{-2pt} 
  \caption{Compression performance with adjusted VoCo token numbers during inference on model trained with fixed numbers.}
\vspace{-8pt} 
  \label{tab:tab7}
\end{table}

\renewcommand{\arraystretch}{0.9} 
\begin{table*}[t]\footnotesize
  \centering
\resizebox{0.9\textwidth}{!}{%
    \begin{tabular}{l|c|c|c|c|c|c|c|c|c|c|c}
    \toprule
      \multirow{2}{*}{Method} & \multirow{2}{*}{Token} & \multicolumn{3}{c|}{\textbf{RefCOCO}}  & \multicolumn{3}{c|}{\textbf{RefCOCO+}} & \multicolumn{2}{c|}{\textbf{RefCOCOg}} & \textbf{GRIT} & \multirow{2}{*}{\textbf{Avg.}} \\
      \cline{3-11}
      & & val & test A & test B & val & test A & test B & val (U) & test (U) & refexp \\
    \midrule
      {\textcolor{darkred}{Upper Bound}} & 256 & 87.01 & 90.61 & 80.24 & 81.60 & 87.36 & 72.12 & 82.27 & 82.19 & 69.34 & 100\% \\
    \midrule
      \multirow{2}{*}{\textbf{VoCo-LLaMA}} & 8 & \textbf{85.17} & \textbf{88.92} & \textbf{79.21} & \textbf{80.02} & \textbf{85.13} & \textbf{70.22} & \textbf{80.36} & \textbf{80.64} & \textbf{68.59} & \textbf{90.7\%} \\
       & 1 &83.29& 86.89& 77.87& 77.62& 83.02& 67.74& 78.32& 78.06& 67.69 & 79.9\%  \\
    \midrule
      {\textcolor{darkgreen}{Lower Bound}} & 1 & 68.34 & 72.96 & 68.03 & 62.58 & 64.77 & 50.65 & 62.30 & 62.99 & 60.50 & 0\% \\
    \bottomrule
    \end{tabular}
    }
\vspace{-2pt} 
  \caption{Compression performance on REC task. Avg. means the average compression retention rate on all benchmarks.}
\vspace{-2pt} 
  \label{tab:tab1}
\end{table*}

\renewcommand{\arraystretch}{0.9} 
\begin{table*}[t]
\scriptsize
  \centering
\resizebox{0.9\textwidth}{!}{%
    \begin{tabular}{l|c|c|c|c|c|c|c|c|c|c}
    \toprule
      \multirow{2}{*}{Method} & \multirow{2}{*}{Token} & \multicolumn{3}{c|}{RefCOCO}  & \multicolumn{3}{c|}{\textbf{RefCOCO+}} & \multicolumn{2}{c|}{RefCOCOg} & \multirow{2}{*}{\textbf{Avg.}} \\
      \cline{3-10}
      & & val & test A & test B & val & test A & test B & val (U) & test (U)  &  \\
    \midrule
      {\textcolor{darkred}{Upper Bound}} & 256 & 75.61 & 44.26 & 104.83 & 56.42 & 40.98 & 68.25 & 62.71 & 65.58 & 100\%  \\
    \midrule
      \multirow{2}{*}{\textbf{VoCo-LLaMA}} & 8 & \textbf{73.87} & \textbf{43.13} & \textbf{102.71} & \textbf{55.34} & \textbf{39.91} & \textbf{67.00} & \textbf{61.59} & \textbf{64.45} & \textbf{91.3\%} \\
      & 1 & 71.92& 41.81& 94.50& 53.98& 38.96& 65.35& 60.46& 63.17 & 81.6\%  \\
    \midrule
      {\textcolor{darkgreen}{Lower Bound}} & 1 & 56.73 & 31.82 & 78.09 & 43.71 & 30.26 & 52.22 & 50.49 & 53.22 & 0\%  \\
    \bottomrule
    \end{tabular}}
\vspace{-2pt} 
  \caption{Compression performance on REG task. Avg. means the average compression retention rate on all benchmarks.}
\vspace{-4pt} 
  \label{tab:tab2}
\end{table*}

We additionally compare our method with previous common learning-based approaches (\emph{i.e.,} Q-Former and average pooling) for vision token compression. 
Our method significantly outperforms previous methods across all benchmarks. 
Specifically, we observe an improvement of 19.6\% in average compression retention rate, surpassing the average pooling compression strategy. 
In contrast, while Q-Former has demonstrated impressive capabilities in capturing visual features with 32 queries, its performance undergoes a substantial decline when the query count is reduced to a single digit.
This proves that our VoCo-LLaMA, which utilizes the knowledge distillation from language models itself, maintains more valuable vision information than that of average pooling or query-based compression.

\noindent\textbf{Number of VoCo tokens.}
We evaluate the impact of the number of VoCo tokens on vision compression performance. 
\cref{tab:voco_tab_2} illustrates the trend of compression performance retention as the number of VoCo tokens varies, where the green and red lines represent the upper and lower bounds of compression performance, respectively. 
We adopted the same training settings and data as in the main experiments.
It can be observed that as the number of VoCo tokens grows, the overall compression performance of the model shows an upward trend. 
Increasing the number of tokens within the range of fewer than 10 tokens results in a significant improvement in compression performance. 
Finally, when conducting 128 VoCo tokens, the model achieves an average compression performance retention rate of 97.7\%, indicating that the performance loss due to compression is almost negligible when compressing into more than 100 tokens.
Interestingly, we observe that when training with 128 VoCo tokens, the result on the SEED-Bench exceeds the performance upper bound model.

\renewcommand{\arraystretch}{0.9}  
\begin{table*}[!htbp]
  \centering
\resizebox{0.99\textwidth}{!}{%
  \footnotesize 
  \begin{tabular}{l|c|c|c@{\hspace{0.5em}}c|c@{\hspace{0.5em}}c|cc}
    \toprule
    Method & Token & KV Cache Length & Storage Memory (MB) & \(\Delta\) & CUDA Time (ms) $\downarrow$  & \(\Delta\) & FLOPs (T) $\downarrow$ & \(\Delta\) \\
    \midrule
    Baseline & 576 & - & - & - & 440.5 & - & 9.6 & -    \\
    Full Caching & 576 & 576 & 302.4 & - & 154.9 & 64.8{\footnotesize \%} & 1.2 & 87.5{\footnotesize \%}     \\
    \midrule
    \textbf{VoCo-LLaMA} & \textbf{1} & \textbf{1} & \textbf{0.525} & {99.8{\footnotesize \%}} & \textbf{134.0} & {69.6{\footnotesize \%}} & \textbf{0.5} & {94.8{\footnotesize \%}}    \\
    \bottomrule
  \end{tabular}
}
\vspace{-2pt} 
  \caption{Efficiency analysis of VoCo-LLaMA including cache storage memory, CUDA time and the FLOPs. \(\Delta\) denotes the reduction ratio.}
\vspace{-2pt} 
\label{tab:voco_tab_4}
\end{table*}

\renewcommand{\arraystretch}{0.9}  
\begin{table*}[!htbp]
\footnotesize
  \centering
\resizebox{0.99\textwidth}{!}{
  \begin{tabular}{lcccc|cccccc}
    \toprule
    \multirow{2}{*}{Method} &  \multirow{2}{*}{Visual Encoder} & \multirow{2}{*}{LLM} & \multirow{2}{*}{Res.} & \multirow{2}{*}{Image Token} & \multicolumn{2}{c}{\textbf{MSVD-QA}} & \multicolumn{2}{c}{\textbf{MSRVTT-QA}} & \multicolumn{2}{c}{\textbf{ActivityNet-QA}} \\
     &  & &  &  & Acc & Score & Acc & Score & Acc & Score \\
    \midrule
    \multicolumn{11}{c}{{\emph{Methods w/o Vision Compression}}} \\
    \midrule
    FrozenBiLM~\cite{yang2022frozenbilm} & CLIP-L & DeVERTa-V2 & 224 & 256 & 32.3 & - & 16.8 & - & 24.7 & -     \\
    Video-LLaMA~\cite{damonlpsg2023videollama} & EVA-G & Vicuna-7B & 224 & 256 & 51.6 & 2.5 & 29.6 & 1.8 & 12.4 & 1.1     \\
    VideoChat~\cite{2023videochat} & - & Vicuna-7B & 224  & - & 56.3 & 2.8 & 45.0 & 2.5 & 26.5 & 2.2     \\
    Video-ChatGPT~\cite{Maaz2023VideoChatGPT} & CLIP-L & Vicuna-7B & 224 & 256 & 64.9 & 3.3 & 49.3 & 2.8 & 35.2 & 2.7     \\
    BT-ADapter~\cite{liu2023btadapter} & CLIP-L & Vicuna-7B & - & - & 67.5 & 3.7 & 57.0 & 3.2 & 45.7 & 3.2     \\
    Vista-LLaMA~\cite{ma2023vistallama} & EVA-G & Vicuna-7B & 224 & 256 & 65.3 & 3.6 & 60.5 & 3.3 & 48.3 & 3.3     \\
    Chat-UniVi~\cite{jin2024chatunivi} & CLIP-L & Vicuna-7B & 224 & 256 & 69.3 & 3.7 & 55.0 & 3.1 & 46.1 & 3.3     \\
    \midrule
    \multicolumn{11}{c}{{\emph{Methods w/ Vision Compression}}} \\
    \midrule
    LLaMA-VID~\cite{li2023llamavid}  & EVA-G & Vicuna-7B & 224 & 2 & 69.7 & 3.7 & 57.7 & 3.2 & 47.4 & 3.3     \\
    \midrule
    \multirow{4}{*}{\textbf{VoCo-LLaMA}}  & \multirow{4}{*}{CLIP-L} & \multirow{4}{*}{Vicuna-7B} & 224 & 2 & \textbf{72.3} & \textbf{3.9} & \textbf{61.1} & \textbf{3.5} & {47.9} & \textbf{3.4}     \\
      &  & & 336 & 2 & \textbf{72.6} & \textbf{3.9} & \textbf{61.2} & \textbf{3.5} & {47.9} & \textbf{3.4}     \\
      &  & & 224 & 8 & \textbf{73.4} & \textbf{3.9} & \textbf{62.0} & \textbf{3.5} & \textbf{48.5} & \textbf{3.4}     \\
      &  & & 336 & 8 & \textbf{73.5} & \textbf{3.9} & \textbf{62.3} & \textbf{3.5} & \textbf{48.6} & \textbf{3.4}     \\
    \bottomrule
  \end{tabular}
  }
\vspace{-2pt} 
  \caption{
  Comparison with leading video understanding methods, with and without vision compression, on three zero-shot benchmarks.}
\vspace{-5pt} 
  \label{tab:voco_vid_1}
\end{table*}

\noindent\textbf{Method of Compression.}  
We compare our method with LLaMA-VID on vision compression, specifically evaluating its full model that utilizes both context and content tokens. 
For a fair comparison, VoCo-LLaMA is trained under the exact same settings and applied the same visual encoder, EVA-G~\cite{fang2023EVA}, in this experiment.
As shown in~\cref{tab:voco_tab_3}, our method outperforms the previous approach when using a single content compression token, even surpassing the performance of LLaMA-VID when it uses multiple context tokens. 
In particular, we could observe an improvement of 2.8 and 3.7 on GQA and VQA$^{T}$ benchmarks, respectively.

\noindent\textbf{Adaptability of VoCo Number.}  
To assess the model's adaptability to varying numbers of compression tokens, we trained the model with a fixed number of tokens and evaluated its performance with different token numbers. 
As demonstrated in~\cref{tab:tab7}, we conducted experiments by fixing the number of VoCo tokens (32) during training and varying the number of tokens during inference. 
Our method achieves better performance with an increasing number of compressed tokens, without requiring specialized training for elastic compressed token.

\noindent\textbf{Results on fine-grained tasks.}  
We analyze the extent of loss of fine-grained visual information after high-magnification compression of vision tokens in our approach. 
Here, we apply our method to~\cite{chen2023shikraunleashingmultimodalllms} which is a cleanly structured MLLM trained on fine-grained task data such as REG, REC, and PointQA. 
As shown in~\cref{tab:tab1} and~\cref{tab:tab2}, when compressing vision tokens to 1 VoCo token, our method maintained an impressive average compression retention rate of 79.9\% and 81.6\% for REC and REG tasks, respectively. 
Furthermore, by increasing the number of VoCo tokens to 8, we observed a significant improvement in the average compression retention rate. 
We observe that VoCo-LLaMA achieves similar compression retention rate to other benchmarks on fine-grained tasks, mainly because the Lower Bound model incurs more information loss on fine-grained tasks.
Please refer to the \textit{supplementary material} for additional fine-grained benchmarks, including VisWiz, OCRBench and others.

\noindent\textbf{Inference Efficiency.}
We discuss the inference efficiency under the scenarios that images are cached as discussed in~\cref{sec:cache}.
Due to our model's design, the representation of compressed image (\emph{i.e.,} transformer activations on top of VoCo tokens) can be stored and repeatedly utilized in the form of a KV cache. 
We conduct a comparative analysis of CUDA time, FLOPs, and KV Cache storage size during the inference process, and compare our method with the baseline method and the full caching method.
The baseline method, as its name suggests, does not employ any caching strategy and directly encodes and infers images. 
In contrast, the full caching method stores the uncompressed Transformer activations upon all vision tokens as KV caches. 
More specifically, we follow the approach of~\cite{pope2022efficiently}, storing the keys and values of each Transformer layer.
As displayed in~\cref{tab:voco_tab_4}, we conduct an inference efficiency analysis on a single NVIDIA A100 using identical lengths of text prompts and single-image inputs. 
Compared to the baseline model without caching strategy, VoCo-LLaMA achieves a significant reduction of 69.6\% in CUDA time and 94.8\% in FLOPs.
Relative to the full caching strategy, our method save 99.8\% of cache storage while achieving lower CUDA time and FLOPs, demonstrating the inference efficiency gains brought by our approach.
Please refer to the \textit{supplementary material} for further discussion and details for inference efficiency.

\noindent\textbf{Video Understanding.}
We further evaluate the performance of VoCo-LLaMA on three widely used video understanding benchmarks, reporting results for input image resolutions of 224 and 336, respectively.
First, we discuss the video understanding methods that utilize vision compression. 
Ensuring fair comparison, we adopted the same compression ratio as previous method~\cite{li2023llamavid}, compressing each video frame into 2 VoCo tokens for training and inference. 
Our method consistently outperforms previous video compression methods across all three benchmarks.
Specifically, on the MSVD-QA and MSRVTT-QA datasets, VoCo-LLaMA achieved accuracies of 72.3\% and 61.1\%, respectively, corresponding to absolute gains of 3.7\% and 5.9\% over the previous best methods. 
Moreover, our method achieves the highest scores of 3.9 and 3.5, respectively.

In comparison to video understanding methods that do not employ vision compression, our approach, which represents each video frame with a mere 2 VoCo tokens, demonstrates strong competitiveness against methods that utilize 256 or more vision tokens per frame.
To further explore the potential of VoCo-LLaMA, we opted to compress video frames into the number of VoCo tokens that exhibited the optimal compression performance within the 0 order of magnitude (\emph{i.e.,} 8 tokens).
Notably, as we increase the number of tokens, our method effectively leverages additional visual information.
We also analyze the performance loss caused by vision compression and evaluate on other video QA benchmarks, as detailed in the \textit{supplementary material}.

%% file: sec/5_Conclusion.tex
\section{Conclusion} \label{sec:conclusion}
In this paper, we propose VoCo-LLaMA, the first approach to compress visual information using LLMs.
By distilling the LLMs' understanding of vision tokens into a compact representation, our method can compress hundreds of vision tokens into a single VoCo token, while minimizing information loss. 
VoCo-LLaMA significantly reduces cache storage and boosts efficiency during the inference stage.
Moreover, our method exhibits promising performance in learning temporal understanding on video data with continuous training.
In summary, our approach offers a promising solution for fully utilise the limited context window of VLMs, making them more efficient for multi-modal applications.